\documentclass[10pt,twocolumn,letterpaper]{article}

\usepackage{3dv}
\usepackage{times}
\usepackage{epsfig}
\usepackage{graphicx}
\usepackage{amsmath}
\usepackage{amssymb}
\usepackage{bm}
\usepackage{color}
\usepackage{caption}
\usepackage[pagebackref=true,breaklinks=true,letterpaper=true,colorlinks,bookmarks=false]{hyperref}

\renewcommand{\vec}[1]{\boldsymbol{#1}}

\newcommand{\smpl}[0]{M}
\newcommand{\posefun}[0]{T}
\newcommand{\blendfun}[0]{W}
\newcommand{\offsetfun}[0]{B}
\newcommand{\jointfun}[0]{J}

\newcommand{\pose}[0]{\boldsymbol{\theta}}
\newcommand{\shape}[0]{\boldsymbol{\beta}}

\newcommand{\blendweights}[0]{\mathbf{W}}

\newcommand{\templatetpose}[0]{\mathbf{T}_\mu}

\newcommand{\offsets}{\mathbf{D}}

\threedvfinalcopy %
\def\threedvPaperID{211} %

\ifthreedvfinal\fi
\begin{document}

\title{360-Degree Textures of People in Clothing from a Single Image}

\author{
Verica Lazova\\
\and
Eldar Insafutdinov \\
\and
Gerard Pons-Moll \\
\and \\
Max Planck Institute for Informatics \\
Saarland Informatics Campus, Germany \\
{\tt\small \{vlazova, eldar, gpons\}@mpi-inf.mpg.de}
}

\twocolumn[{
  \renewcommand\twocolumn[1][]{#1}
  \maketitle
  \begin{center}
  \newcommand{\teaserwidth}{1.\textwidth}
  \vspace{-0.3in}
  \centerline{\includegraphics[width=\teaserwidth,clip]{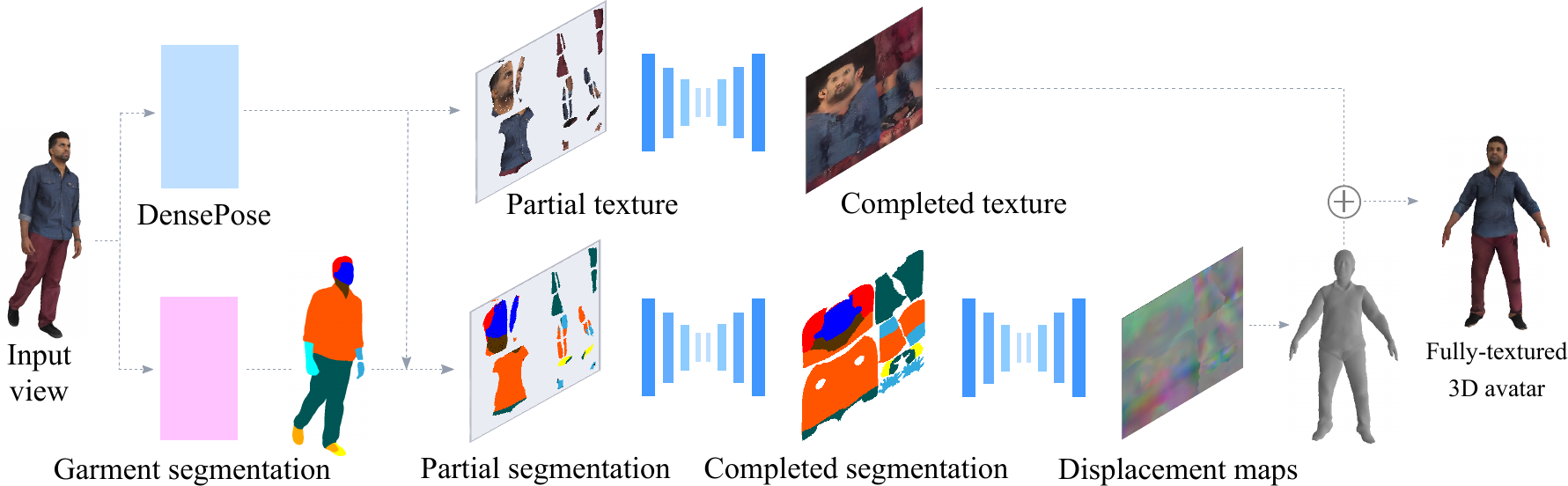}}
  \captionof{figure}{Given a single view of a person we predict a complete texture map in the UV space, complete clothing segmentation as well as a displacement map for the SMPL model \cite{smpl2015loper}, which we then combine to obtain a fully-textured 3D avatar.}
  \label{fig:teaser}
  \end{center}
}]
\maketitle

\begin{abstract}
In this paper we predict a full 3D avatar of a person from a single image. We infer texture and geometry in the UV-space of the SMPL model using an image-to-image translation method. Given partial texture and segmentation layout maps derived from the input view, our model predicts the complete segmentation map, the complete texture map, and a displacement map. The predicted maps can be applied to the SMPL model in order to naturally generalize to novel poses, shapes, and even new clothing. In order to learn our model in a common UV-space, we non-rigidly register the SMPL model to thousands of 3D scans, effectively encoding textures and geometries as images in correspondence. This turns a difficult 3D inference task into a simpler image-to-image translation one. Results on rendered scans of people and images from the DeepFashion dataset demonstrate that our method can reconstruct plausible 3D avatars from a single image. We further use our model to digitally change pose, shape, swap garments between people and edit clothing. To encourage research in this direction we will make the source code available for research purpose~\cite{rvh}. 
\end{abstract}

\section{Introduction}
\label{sec:introduction}
3D models of humans, comprising personalized surface geometry, and full \emph{texture} are required for numerous applications including VR/AR, gaming, entertainment or human tracking for surveillance. Methods capable of recovering such 3D models from a single image would democratize the acquisition process, and allow people to easily recover avatars of themselves. 

While there is extensive work on 3D human pose and shape recovery (surface geometry) from a single image, almost no work addresses the problem of predicting a complete texture of a person from a single image. We, humans, can do it to some degree, because we can guess what the person might look like from another view. We make these guesses effortlessly because we have seen people from many angles, and we have built a mental model of typical correlations. For example, the texture patterns within a garment are often repetitive; skin color and hair is roughly homogeneous; and the appearance of the left and right shoe is typically the same.  

\begin{figure}
\includegraphics[angle=-90,width=0.48\textwidth]{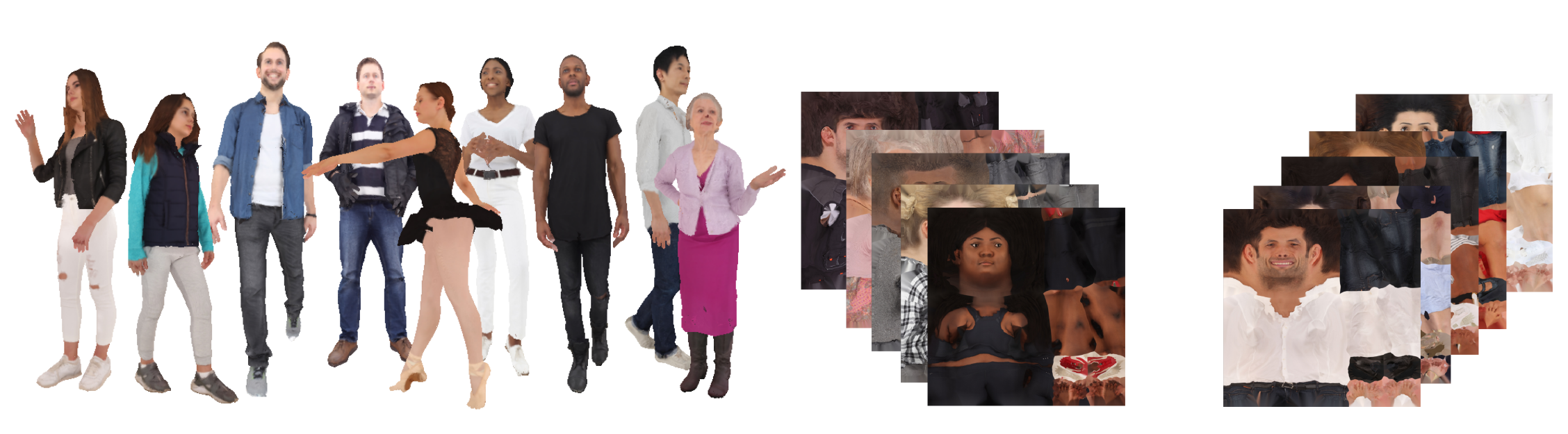}
\caption{\emph{360$^{\circ}$ People Textures}: Example registrations and texture maps used to train our model. After registering SMPL to the scans their appearances are encoded as texture maps in a common UV-space.}
\label{fig:3Dpeople}
\end{figure}

\begin{figure}
\centering
\includegraphics[width=0.47\textwidth]{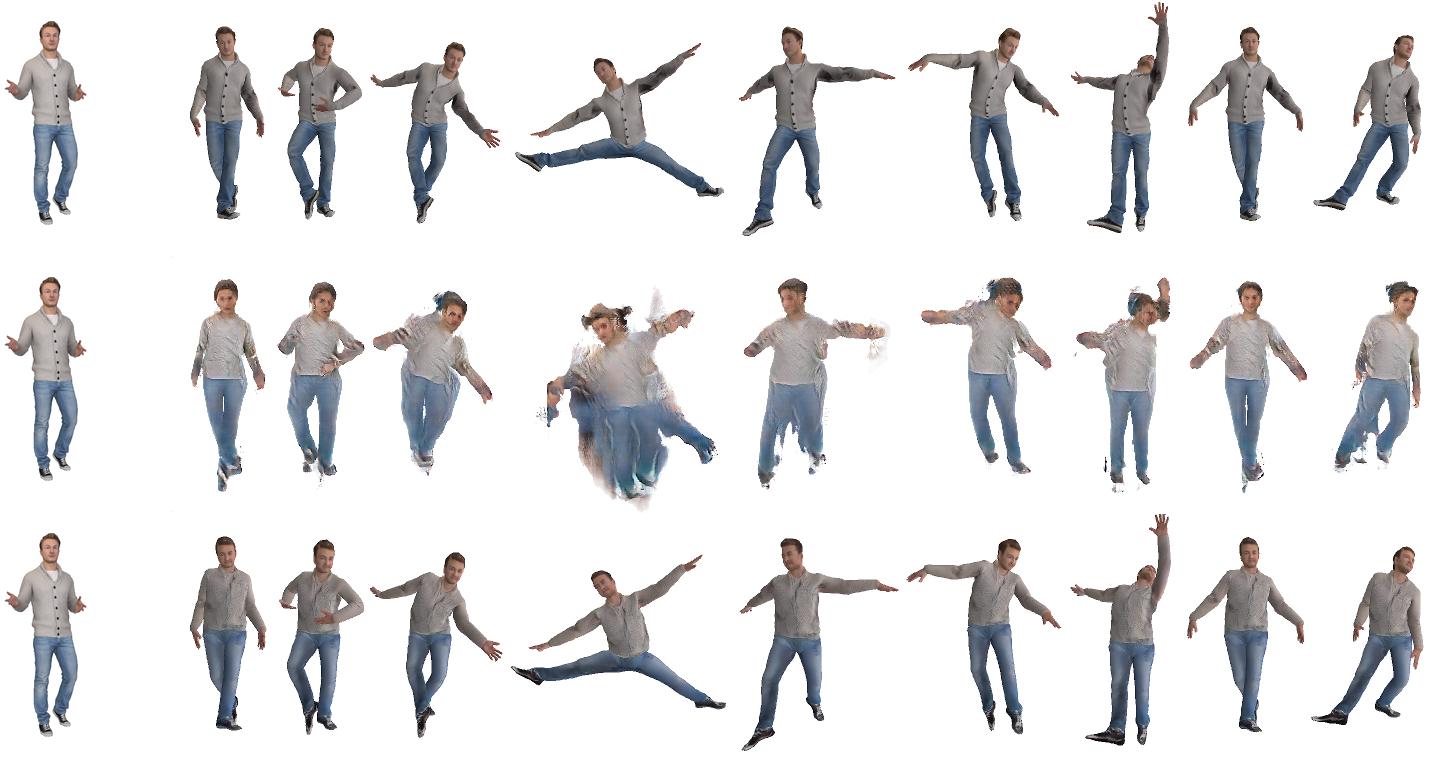}
\caption{Image-based reposing methods cannot handle challenging target poses outside of the training set. First row shows the original scan reposed. The second row shows image based reposing using \cite{siarohin2018deformable}. In the last row is the full model obtained with our method, reposed. The input image is the first one in each row.}
\label{fig:image_based}
\end{figure}

In this work, we introduce a model that automatically learns such correlations from real appearances of people. Specifically, our model predicts a full (360$^{\circ}$) texture map of the person in the UV space of the SMPL model~\cite{smpl2015loper} given a partial texture map derived from the visible part in an image, see Fig.~\ref{fig:teaser}. Instead of learning an image-based model that has to generalize to every possible pose, shape and viewpoint~\cite{han2018viton,Lassner:GP:2017,ma2017pose}, our idea is to learn to complete the full 3D texture and generalize to new poses, shapes and viewpoints with SMPL and traditional rendering (Figure~\ref{fig:image_based}).
Learning in a common UV-space requires texture maps of people in clothing in good correspondence, which is a highly non-trivial task. To that end, we non-rigidly deform the SMPL body model to 4541 static 3D scans of people in different poses and clothing. 
This effectively brings the 3D appearances of people into a common 2D UV-map representation, see Fig.~\ref{fig:3Dpeople}. Learning from this data has several advantages. First, since every pixel in a UV-map corresponds to one location on the body surface, it is easier for a neural network to implicitly learn localized and specialized models for particular parts of the body surface. For example, our model learns to complete different textures for the different garments, the face region and the skin, with clearly defined boundaries. 
Second, at test time, once the texture map is recovered, we can apply it to SMPL and animate it directly in 3D with any pose and shape and viewpoint to generate a coherent movement and appearance--without the risk of failing to generalize to these factors.

Additionally, our model can predict, on the SMPL UV map, a full clothing segmentation layout, from which we predict a \emph{plausible} clothing geometry of the garments worn by the subject. The latter is a highly multi-modal problem where many clothing geometries can explain the same segmentation layout. However, as observed for other tasks~\cite{isola2017image} an image translation network is capable of producing a plausible output. 
To allow for further control, we can additionally easily edit the predicted segmentation layout in order to control the shape and appearance of the clothing. In particular, we can modify the length of the sleeves in shirts and t-shirts, and interpolate from shorts to pants and the other way around.

We train and our method on our newly created dataset (\emph{360$^{\circ}$ People Textures}) consisting of 4541 scans registered to a common SMPL body template. Our results generalize on real images and show that our model is capable of completing textures of people, often reproducing clothing boundaries and completing garment textures and skin. In summary, our model can reconstruct a full 3D animatable avatar from a single image, including texture, geometry and segmentation layout for further control.

\section{Related work}
\label{sec:related}
{\bf Pose guided image and video generation}
Given a source image and a target 2D pose, image-based methods~\cite{ma2017pose,ma2018disentangled,esser2018variational,si2018multistage,neverova2018dense} produce an image with the source appearance in the target pose.  
To deal with pixel miss-alignments, it is helpful to transform the pixels in the original image to match the target pose within the network~\cite{siarohin2018deformable,balakrishnan2018synthesizing,lorenz2019unsupervised}.  Following similar ideas to growing GANs~\cite{karras2017progressive}, high resolution anime characters can be generated~\cite{hamada2018full}. Dis-occlusions and out of plane rotations are a problem for such methods.

Sharing our goal of incorporating more information about the human body during the generation process, some works~\cite{Lassner:GP:2017,neverova2018dense} also leverage the SMPL body model. By conditioning on a posed SMPL rendering, realistic images of people in the same pose can be generated~\cite{Lassner:GP:2017}--the model however cannot maintain the appearance of the person when the pose changes. Recent approaches~\cite{neverova2018dense,grigorev2018coordinate} leverage DensePose~\cite{DensePose} to map a source appearance to the SMPL surface, and condition the generated image on the target DensePose mask. Although their goal is generating an image, the model generates a texture map on the SMPL surface as an internal representation.  During training, multiple views of the same subject are mapped to a custom designed UV-map, and the network is forced to complete the texture on the visible parts. 
While training from images is practical, DensePose was not designed to accommodate for hair and clothing deviating from the body, and significant miss-alignments might occur in the UV-map when DensePose fails or is inaccurate. Consequently, texture map completion results are limited by the ability of DensePose to parse images during training.
We use DensePose solely to create the partial texture map from the input view, but train from high quality aligned and complete texture maps, which are obtained by registering SMPL to 3D scans. This has several advantages. By virtue of the registration, the textures are very well aligned in the SMPL-UV space, and cover the full extend of the appearance including clothing and hair. This allows us to generalize beyond poses seen in a particular dataset such as DeepFashion~\cite{liu2016deepfashion}. Furthermore, we go beyond~\cite{neverova2018dense,grigorev2018coordinate} and predict a full 3D textured avatar including geometry and 3D segmentation layout for further control. 

While image-based methods are able to generate plausible images, many applications (VR/AR, gaming) require a full 3D representation of the person. Traditional graphics rendering could potentially be replaced by querying such models for every desired novel viewpoint and pose. Unfortunately, image-based models do not produce temporally coherent results for sequences of target poses. Recent work focus on video generation~\cite{chan2018everybody,balakrishnan2018synthesizing,yang2018pose,shysheya2019textured}, but these models are often trained specifically for a particular subject. Furthermore, the formation of more complex scenes with objects or multiple people would be a challenge for image-based methods. 

{\bf Try-on and conditional clothing synthesis} A growing number of recent works focus on conditioning the person image generation on a particular clothing item~\cite{han2018viton,wang2018toward}, or a description~\cite{zhu2017be}. Other works demonstrated models capable of swapping the appearances between two different subjects~\cite{raj2018swapnet,Zanfir_2018_CVPR}. \\
{\bf Multi-view texture generation.} 
Texture generation is challenging even in the case of multi-view images. The difficulty lies in combining the partial textures created from different views by using blending~\cite{bernardini2001high, debevec1996modeling, Ofek:1997:MTI:616045.618417,starck2003model}, mosaicing~\cite{baumberg2002blending,lensch2001silhouette,niem1997automatic,rocchini1999multiple}, graph cuts~\cite{lempitsky2007seamless}, or flow based registration~\cite{eisemann2008floating,bi2017patch,waechter2014let,FuYanYangEtAl,zhou2014color,alldieck2018detailed} in order to reduce ghosting and stitching artifacts. 
A learning based model that leverages multi-view images for training~\cite{neverova2018dense,grigorev2018coordinate} will suffer similar problems. Hence, in this work, we learn from complete texture maps obtained from 3D registrations.\\
{\bf 3D person reconstruction from images}
While promising, recent methods for 3D person reconstruction either require video as input~\cite{alldieck19cvpr,alldieck2018detailed,alldieck2018video}, scans~\cite{yang2018analyzing}, do not allow control over pose, shape and clothing~\cite{natsume2018siclope,saito2019pifu}, focus only on faces~\cite{yamaguchi2018high,huynh2018mesoscopic,tewari2018self,sela2017unrestricted,nagano2018pagan,tewari2018fml}, or only on garments \cite{wang2018learning}.

\section{Method}
\label{sec:method}
We introduce an image-to-image translation method for inferring texture and geometry of a person from one image. Our method predicts a full texture map, displacement map and full clothing segmentation for additional control. Our models are trained on large number of high quality, diverse 3D scans of people in clothing.
\subsection{Synthetic data generation}
To generate our training set, we take 4541 static scans, and register non-rigidly the SMPL template to them to obtain complete textures aligned in the SMPL UV space. We will briefly review SMPL and explain our non-rigid registration procedure in~\ref{subsec:registration}.
\subsubsection{SMPL Body Model with Clothing}
\label{subsec:SMPL}
\begin{figure}
\centering
\includegraphics[angle=-90,width=0.48\textwidth]{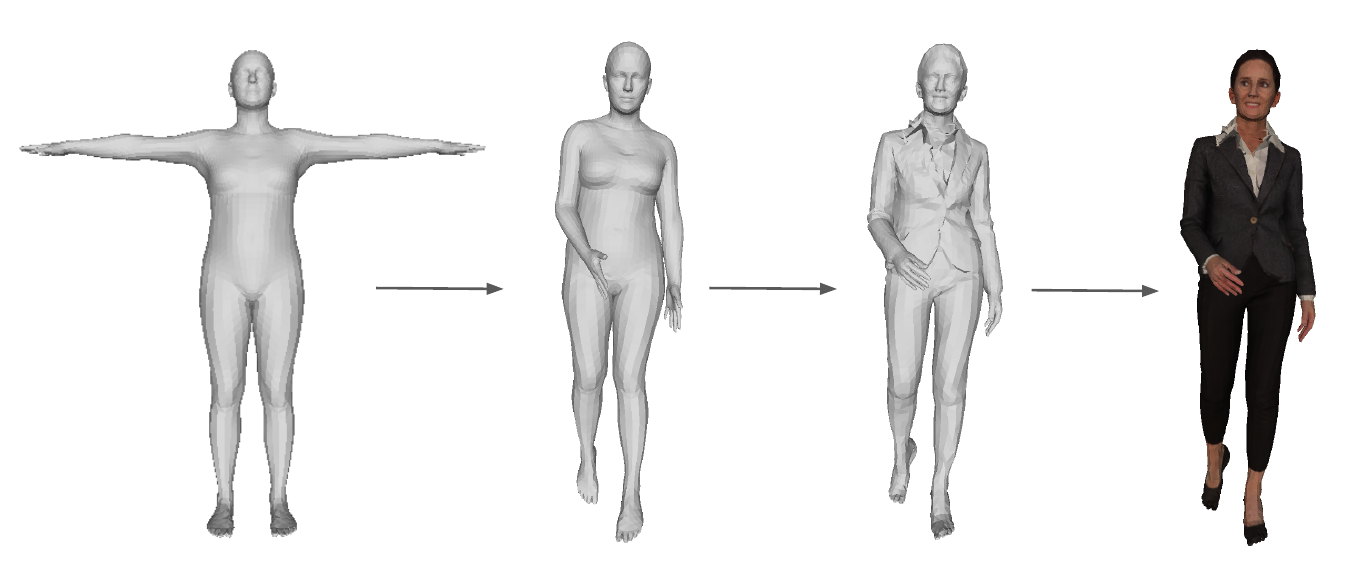}
\caption{Registration: we bring all scan appearances into correspondence by non-rigidly deforming the SMPL model. This allows us to learn in a pose invariant space.}
\label{fig:alignment}
\end{figure}

SMPL is a function $M(\pose,\shape)$ that takes as input pose $\pose\in\mathbb{R}^{72}$ and shape parameters $\shape \in \mathbb{R}^{10}$ and outputs the $N=6890$ vertices of a mesh. SMPL applies deformations to the mean shape $\templatetpose$ using a learned skinning function:
\begin{equation}
\smpl(\shape,\pose) = \blendfun(\posefun(\shape,\pose), \jointfun(\shape), \pose, \blendweights)
\end{equation}
\begin{equation}
\posefun(\shape,\pose) = \templatetpose + \offsetfun_s(\shape) + \offsetfun_p(\pose),
\end{equation}
where $\blendfun$ is a linear blend-skinning function 
applied to the morphed shape $\posefun(\shape,\pose)$ based on the skeleton joints $\jointfun(\shape)$; the morphed shape is obtained by applying pose-dependent deformations $\offsetfun_p(\pose)$ and shape-dependent deformations $\offsetfun_s(\shape)$. The shape space $\offsetfun_s(\shape)$ was learned from undressed scans of people and therefore cannot accommodate for clothing. Since our goal is registering the SMPL surface to scans of people in clothing, we modify SMPL by adding a set of offsets $\offsets \in \mathbb{R}^{3N}$ to the template:
\begin{equation}
\smpl(\pose, \shape, \offsets) = \blendfun(\posefun(\pose, \shape, \offsets), \jointfun(\shape), \pose, \blendweights)
\end{equation}
\begin{equation}
\posefun(\pose,\shape,\offsets) = \templatetpose + \offsetfun_s(\shape) + \offsetfun_p(\pose)+\offsets,
\label{eq:offset_SMPL}
\end{equation}
which are responsible to explain clothing, hair and details beyond the shape space of SMPL. 
\paragraph{Non-rigid registration}
\label{subsec:registration} of humans (Figure \ref{fig:alignment}) is challenging even for scans without clothing~\cite{dfaust:CVPR:2017,pons2015dyna}. The scans in \emph{360$^{\circ}$ People Textures} include subjects in clothing scanned in variety of poses. 
Registration without a good initialization of pose and shape would fail. To obtain a good initialization, we render the scans in $8$ camera viewpoints around the vertical axis and run a 2D pose joint detector~\cite{cao2017realtime} in each of the views. Then, we minimize the re-projection error between SMPL joint positions $J(\pose,\shape)$ and the detected joints $\mathbf{j}^i_{2D}$ with respect to pose and shape:
\begin{align}
\arg\min_{\pose,\shape} \sum_{i} \|\pi_{C_i}(J(\pose,\shape))- \mathbf{j}^i_{2D}\|_F^2 + \\
\lambda_{\pose} D_M(\pose;\boldsymbol{\mu}_{\pose},\boldsymbol{\Sigma}_{\pose}) + \lambda_{\shape} D_M(\shape;\boldsymbol{\mu}_{\shape},\boldsymbol{\Sigma}_{\shape}),
\end{align}
where  $\pi_{C_i}$ projects the 3D joints $J(\pose,\shape)$ onto the view of camera $C_i$, and $D_M(\dot)$ is a Mahalanobis distance prior that we use to regularize pose and shape. 
This brings the model close to the scan, but it is not registered at this point because SMPL cannot explain details such as clothing, hair or jewellery.
Let $\mathcal{S}$ denote a scan we want to register to, and $\mathcal{A} = \{\vec{A},\vec{F}\}$ be the registration mesh with free-form vertices $\mathbf{A}\in \mathbb{R}^{3N}$ and faces $\mathbf{F}$ defining the same topology as SMPL. We obtain a registration by minimizing the following objective function:
\begin{align}
\small
\arg\min_{\pose,\shape,\mathbf{A}} \sum_{\vec{x}_s \in \mathcal{S}} \rho(\operatorname{dist}(\mathcal{A},\vec{x}_s)) +  \sum_{i=0}^{N} w_i \|\mathbf{A}_i - M_i(\pose,\shape)\|^2_F + \nonumber\\
+\lambda_{\pose} D_M(\pose;\boldsymbol{\mu}_{\pose},\boldsymbol{\Sigma}_{\pose}) + \lambda_{\shape} D_M(\shape;\boldsymbol{\mu}_{\shape},\boldsymbol{\Sigma}_{\shape}),
\label{eq:registration}
\end{align}
where the first term measures the distance from every point in the scan $\mathbf{x}_s \in \mathcal{S}$ to the surface of the registration $\mathcal{A}$, the second term forces the registration vertices $\mathbf{A}_i$ to remain close to the model vertices $M(\pose,\shape)_i$, and the other terms are the aforementioned priors. Here, $\rho(\dot)$ denotes a Geman-McClure robust cost function, and $w_i$ are the weights that penalize deviations from the model more heavily for the vertices on the hands and the feet. The first term in Eq.~\eqref{eq:registration} pulls the registration to the scan, while the second penalizes deviations from the body model ensuring the registration looks like a human. That allows to accurately bring all scans into correspondence. 

\subsection{3D Model Generation}
Our full pipeline for 3D model generation from an image is shown in Figure \ref{fig:teaser} and consists of three stages: texture completion, segmentation completion and geometry prediction. Given an image of a person and it's garment segmentation we extract partial texture map as well as partial segmentation map. We have put together inpainting models to complete the garment segmentation and the texture in UV-space. The texture completion part recovers the full appearance of the person in the image. The completed segmentation is further used to predict the geometry. To obtain geometry details we predict a displacement map from the segmentation, which gives us a set of offsets from the average SMPL \cite{smpl2015loper} body model, that correspond to clothing geometry. All three parts are trained separately and allow us to go from a single image to a 3D model. Additionally, the complete segmentation map allows us to edit the final model by changing the garment properties such as length or texture, or to completely redress the person.  

\subsubsection{Texture Completion}
The texture completion part recovers the full appearance of the person in the image. We use DensePose \cite{DensePose} to find pixel correspondences between the image and the SMPL body model, which we use to remap a partial texture map of the person in the SMPL texture template (Figure \ref{fig:teaser}). We use an inpainting network based on image-to-image translation \cite{isola2017image,choi2018stargan,wang2018high} that learns to complete the partial texture map. The inaccuracies of texture extraction with DensePose introduce distortions in the partial texture map. As a result, the input and the output image for the texture completion network are not in perfect alignment. 
This makes our task different from the classical image inpainting, where the network can learn to copy the visible parts of the image to the target. 
We use an architecture for the generator based on residual blocks similar to \cite{choi2018stargan, zhu2017unpaired, wang2018high}. We refer the reader to the supplementary material for the detailed description of the network layers.

We train the inpainting network using the conditional Generative Adversarial Network (GAN) approach with a fully-convolutional patch discriminator to discriminate between real and fake texture maps. 
The partial texture map $x$ is fed through the generator $G$ to infer the full texture map $G(x)$. The conditional discriminator $D$ takes pair of partial and complete texture map and classifies them as real or fake, i.e. $D(x, y) = 1$ and $D(x, G(x)) = 0$. We train the generator $G$ by optimizing the compound objective $\mathit{L}_{\mathrm{tex}}$ consisting of a GAN, reconstruction, perceptual and dissimilarity term.
\begin{equation}
\mathit{L}_{\mathrm{tex}} = \lambda_1 \mathit{L}_{\mathrm{GAN}} + \lambda_2\mathit{L}_{\mathrm{recon}} + \lambda_3\mathit{L}_{\mathrm{perc}} + \lambda_4\mathit{L}_{\mathrm{DSSIM}}
\label{eq:full_loss}
\end{equation}
where $\lambda_1$, ..., $\lambda_4$ are the weights for each of the terms.
We use the Wasserstein GAN loss \cite{arjovsky2017wasserstein,gulrajani2017improved} as our adversarial loss, which produced visually sharper results:
\begin{equation}
\mathit{L}_{\mathrm{GAN}} = \mathbb{E}_{x, y} [D(x, y)] - \mathbb{E}_{x} [D(x, G(x))]
\label{eq:wgan_loss}
\end{equation}
The reconstruction loss is an $L1$ distance between the target $y$ and the reconstructed texture map $G(x)$.
\begin{equation}
\mathit{L}_{\mathrm{recon}}(G) = \mathbb{E}_{x, y} ||G(x) - y||_1
\label{eq:recon}
\end{equation}
Additionally we use perceptual loss which involves matching deep features between the target and the reconstructed image. The features are extracted from the convolutional layers of a pretrained VGG-19 neural network, $VGG_l(y)$ and $VGG_l(G(x))$ for each layer $l$. The perceptual loss is the $L1$ distance between features extracted from the target and the generated texture image.
\begin{equation}
\mathit{L}_{\mathrm{perc}}(G) = \mathbb{E}_{x, y} \sum_l ||VGG_l(y) - VGG_l(G(x))||_1
\label{eq:perceptual}
\end{equation}
Finally, as another perceptual term we minimize the dissimilarity index between the target and generated texture map:
\begin{equation}
\mathit{L}_{\mathrm{DSSIM}}(G) = \frac{1 - MSSIM(y, G(x))}{2},
\label{eq:dssim}
\end{equation}
where $MSSIM$ is a Multiscale Structural Similarity index \cite{wang2004image, wang2003multiscale}, a structural similarity computed at 5 different scales.
Please check the supplementary material for more details on the effect of each of the loss functions.

\subsubsection{Segmentation completion}
Similar to the texture completion stage, the segmentation completion starts with a partial clothing segmentation obtained from a single image, and the task is to obtain a valid full body clothing segmentation in UV space. An image-to-image translation model is trained to inpaint the missing part of the segmentation. The architecture for generator and discriminator is the same as in the texture completion network. We train the segmentation completion network by minimizing the $\mathit{L}_{\mathrm{seg}}$ loss:
\begin{equation}
\mathit{L}_{\mathrm{seg}} = \lambda_1 \mathit{L}_{\mathrm{GAN}} + \lambda_2\mathit{L}_{\mathrm{recon}} 
\label{eq:full_loss_seg}
\end{equation}
where $\mathit{L}_{\mathrm{GAN}}$ and $\mathit{L}_{\mathrm{recon}}$ are defined in Equations \ref{eq:wgan_loss} and \ref{eq:recon}. Finally, the inpainted segmentation is discretized by assigning the label for each pixel in a nearest-neighbour fashion.

\subsubsection{Displacement map prediction}
The final step of our pipeline is geometry prediction where the purpose is to generate geometry that corresponds to the clothes the person is wearing. For this purpose, to capture the clothing shape, we learn to generate vertex offsets from the average SMPL body conditioned on garment type, i.e the segmentation. The offsets are stored as displacement map, where each pixel location corresponds to a point on the SMPL model, and each pixel stores the normalized offset at that body location. This allows us to treat the displacements as images and apply the image-to-image translation method we described previously.
A similar approach is used in \cite{alldieck2019tex2shape}, where the authors learn full displacement and normal maps from a partial texture. Our intent is to learn a model which is able to sample plausible geometry that fits the clothing segmentation. The predicted geometry might not exactly coincide with the image (eg. it is impossible to know from a segmentation in UV-space, if a T-shirt is tight or loose). Nevertheless, relying on the segmentation has advantages: it provides additional flexibility, making editing and adjusting the final 3D model straightforward. We discuss some of the model editing possibilities in Section \ref{sec:experiments}.

As in the previous stages, the model for displacements prediction consists of a generator $G$, that takes a segmentation map as input and produces a displacement map at the output. The generator and the discriminator have the same architecture as in the texture completion scenario. We train the model by minimizing the $\mathit{L}_{\mathrm{disp}}$ objective:
\begin{equation}
\mathit{L}_{\mathrm{disp}} = \lambda_1 \mathit{L}_{\mathrm{GAN}} + \lambda_2\mathit{L}_{\mathrm{recon}}  + \lambda_4\mathit{L}_{\mathrm{DSSIM}}
\label{eq:full_loss_disp}
\end{equation}
where $\mathit{L}_{\mathrm{GAN}}$, $\mathit{L}_{\mathrm{recon}}$ and $\mathit{L}_{\mathrm{DSSIM}}$ are defined in Eq. \ref{eq:wgan_loss}, \ref{eq:recon} and \ref{eq:dssim}.

The generated displacement map applied to the average SMPL body yields the full untextured 3D model. Applying the generated texture from the texture completion part on the model, gives us the full 3D avatar (Figure \ref{fig:teaser})

\section{Experiments and Results}
\label{sec:experiments}
In the following section we give more details on our \emph{360$^{\circ}$ People Textures} dataset and we describe our experiments. For details regarding the training process please check the suplementary material. We have made an attempt for quantitative evaluations, however the typical metrics for evaluating GAN models known in the literature did not correspond to the human-perceived quality of the generated texture. Therefore, here we only present our qualitative results.

\begin{figure*}[!h]
\centering
\includegraphics[width=0.96\textwidth]{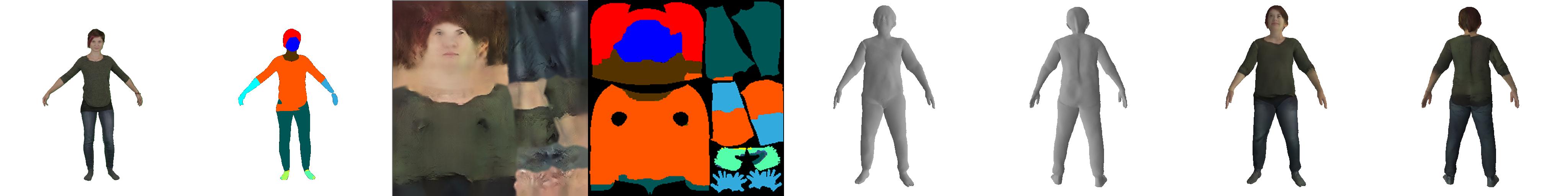}
\includegraphics[width=0.96\textwidth]{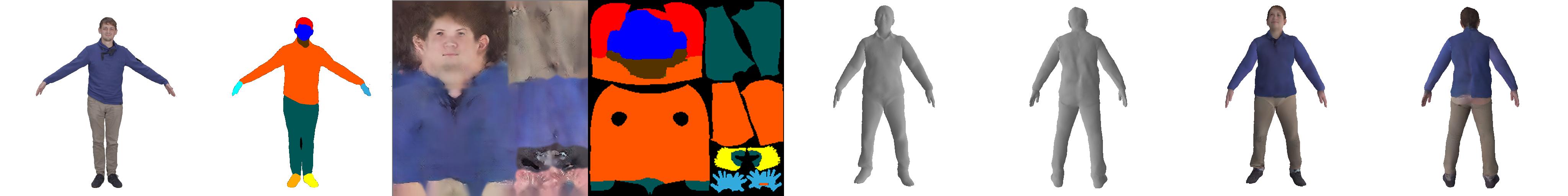}
\includegraphics[width=0.96\textwidth]{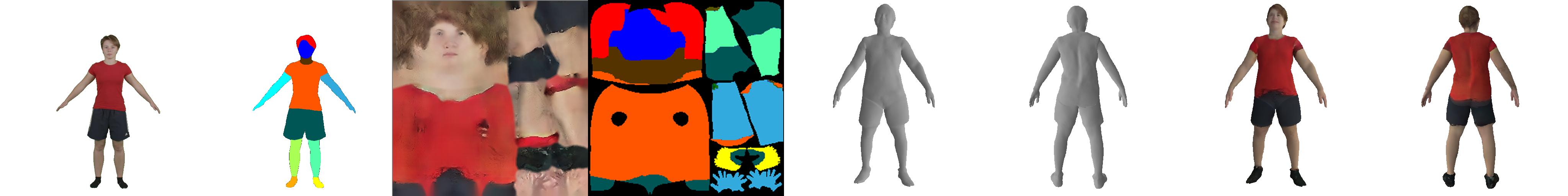}
\includegraphics[width=0.96\textwidth]{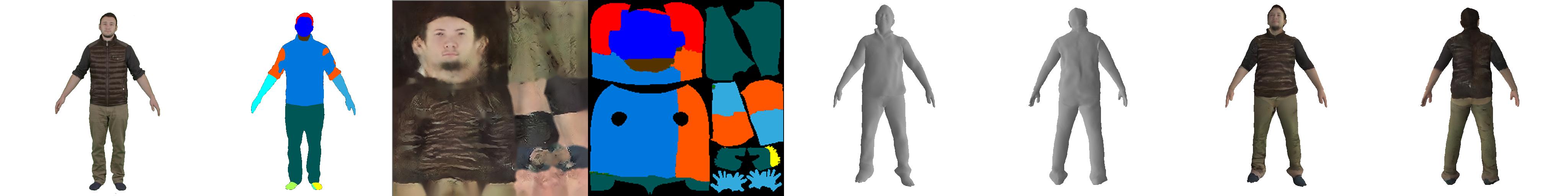}
\includegraphics[width=0.96\textwidth]{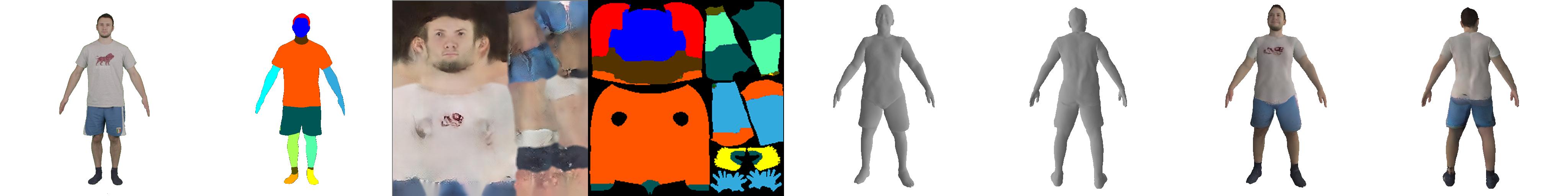}
\includegraphics[width=0.96\textwidth]{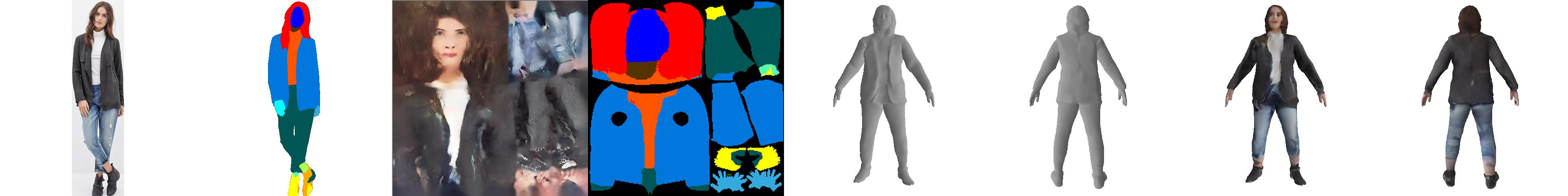}
\includegraphics[width=0.96\textwidth]{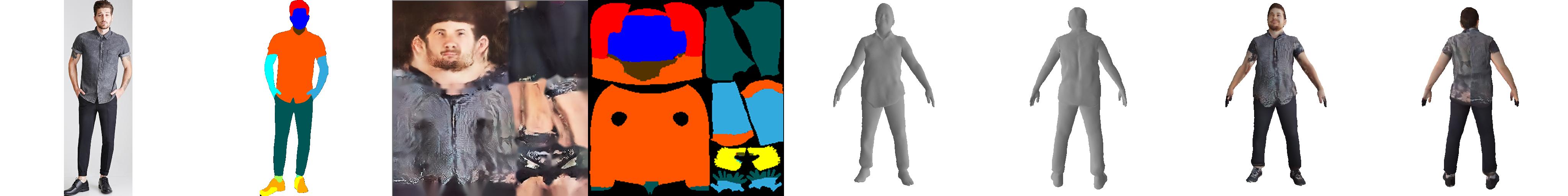}
\includegraphics[width=0.96\textwidth]{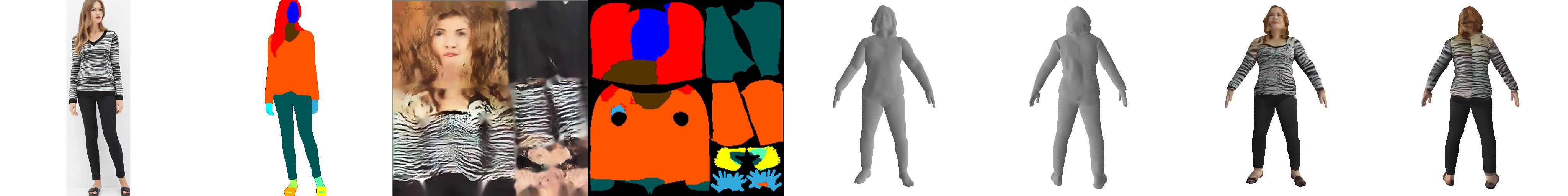}
\includegraphics[width=0.96\textwidth]{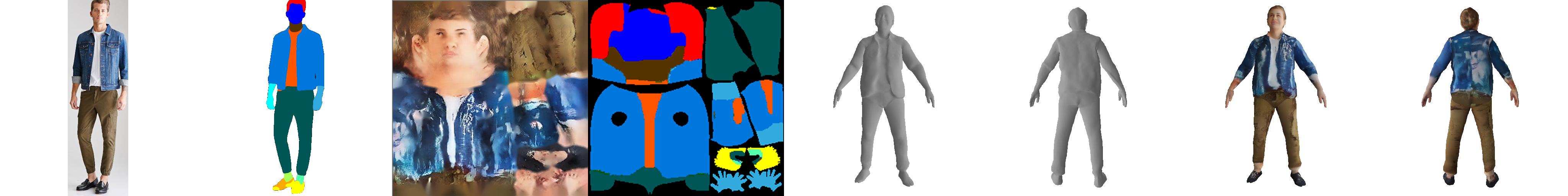}
\includegraphics[width=0.96\textwidth]{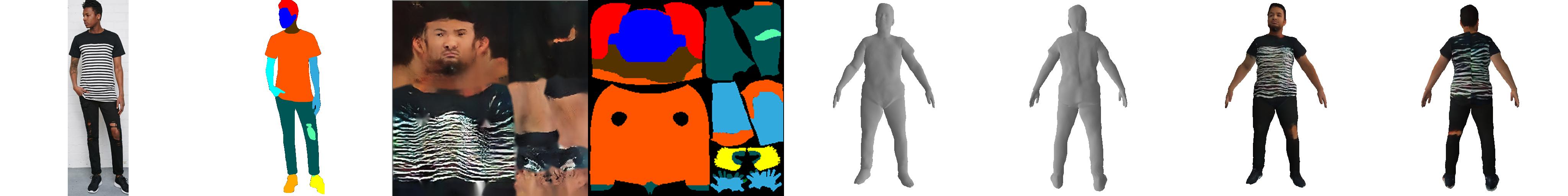}
\caption{Generalization of our method to real images. Left to right: real image, segmented image, complete texture map, complete segmentation map, generated untextured avatar, generated textured avatar.}
\label{fig:results}
\end{figure*}

\begin{figure*}[!t]
\centering
\includegraphics[width=\textwidth]{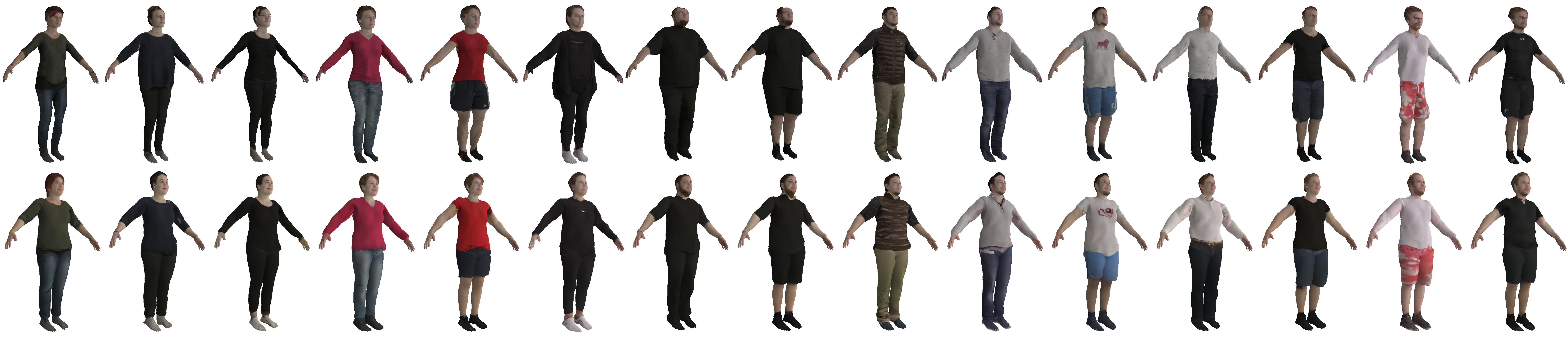}
\caption{Comparison of our single view method (bottom row) to the monocular video method of \cite{alldieck19cvpr} (top row)}
\label{fig:comparison_alldieck}
\end{figure*}

\subsection{Dataset} 
We trained our models on a dataset of 4541 3D scans: 230 come from Renderpeople \cite{renderpeople}; 116 from AXYZ \cite{axyz}; 45 are courtesy of Treedy's \cite{treedys} and the rest 4150 are courtesy of Twindom \cite{twindom}. We used all of them to train the texture inpainting part, plus additional 242 texture maps from the SURREAL \cite{varol17_surreal} dataset. Each of the scans is rendered from 10 roughly frontal views, with the random horizontal rotation of the body sampled from the range $[-\frac{\pi}{2}, \frac{\pi}{2}]$. Out of these we used 40 scans for validation and 10 for testing. Additionally we test on the People Snapshot Dataset \cite{alldieck2018video} and on  preselected 526 images from the DeepFashion \cite{liu2016deepfashion} dataset that satisfy the criteria of fully visible person from a roughly frontal view without overly complex clothing or poses. We use DensePose to extract $256 \times 256$ partial texture maps which we aim to complete.

In order to obtain a full segmentation map first we render the scan from 20 different views. For each of the rendered images we run the clothing segmentation method in \cite{gong2018instance}. Each of the segmented images is projected back to the scan and remapped into a partial segmentation UV-map. All the partial segmentation maps are then stitched together using a graph-cut based method to finally obtain the full segmentation map. The segmentation maps obtained in this way are used as ground-truth for the segmentation completion method. The partial segmentation maps used as input are extracted with DensePose.  

For the geometry prediction part we have selected a subset of 2056 scans for which we could obtain good ground truth segmentation and displacement maps. Out of these 30 scans were taken out for validation. We obtain the segmentation as explained above and the ground truth displacement maps are obtained directly from the scan registrations.

\subsection{Results}
Figure \ref{fig:results} shows sample results of going from real images to 3D models on the People Snapshot and DeepFashion datasets. Each row shows the original image, it's garment segmentation, the recovered full texture and segmentation, predicted geometry and the fully textured avatar. Predicting the body shape is not part of our contribution, but an off-the-shelf method such as \cite{omran2018neural,hmrKanazawa17,bogo2016smplify} could easily be deployed to obtain a more faithful representation of the body shape. Additionally, in Figure~\ref{fig:comparison_alldieck} we show comparison of our single-view method to the monocular video method of \cite{alldieck19cvpr}. While our results look comparable and visually pleasing, should be noted that the method in \cite{alldieck19cvpr} produces higher resolution reconstructions. The advantage of our method is that it requires just a single view.

\begin{figure}[h!]
\centering
\includegraphics[width=0.47\textwidth]{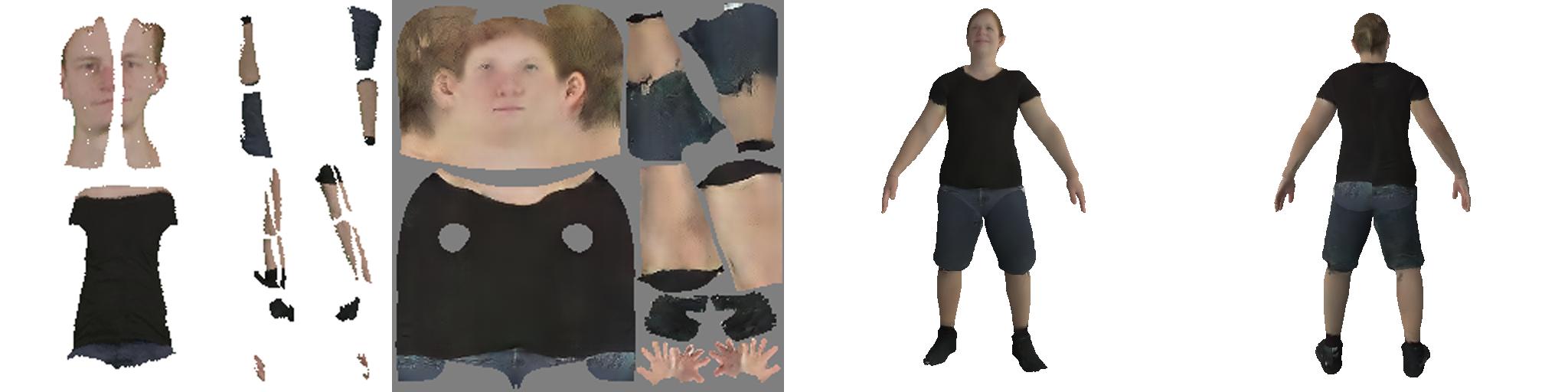}
\includegraphics[width=0.47\textwidth]{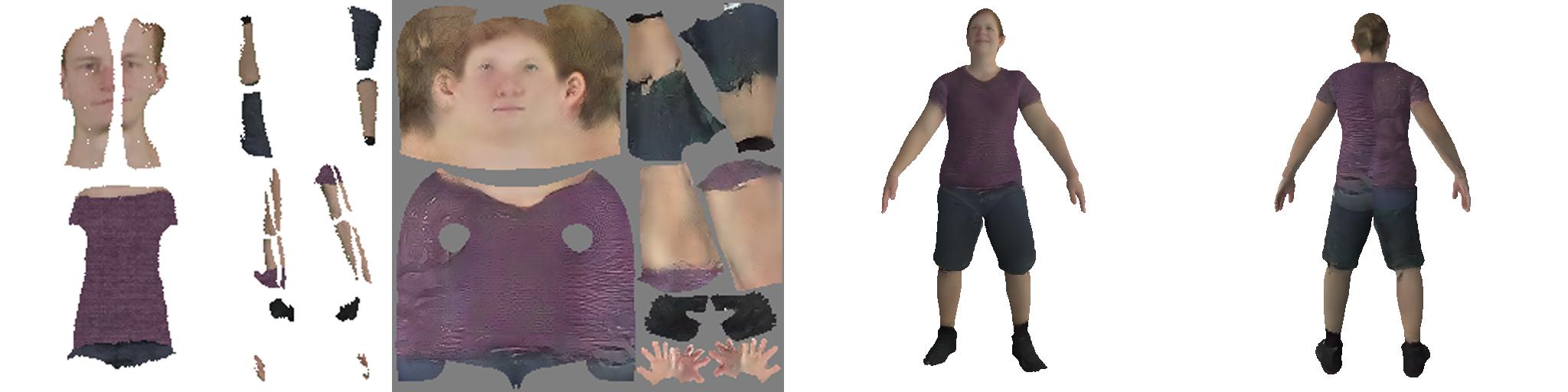}
\includegraphics[width=0.47\textwidth]{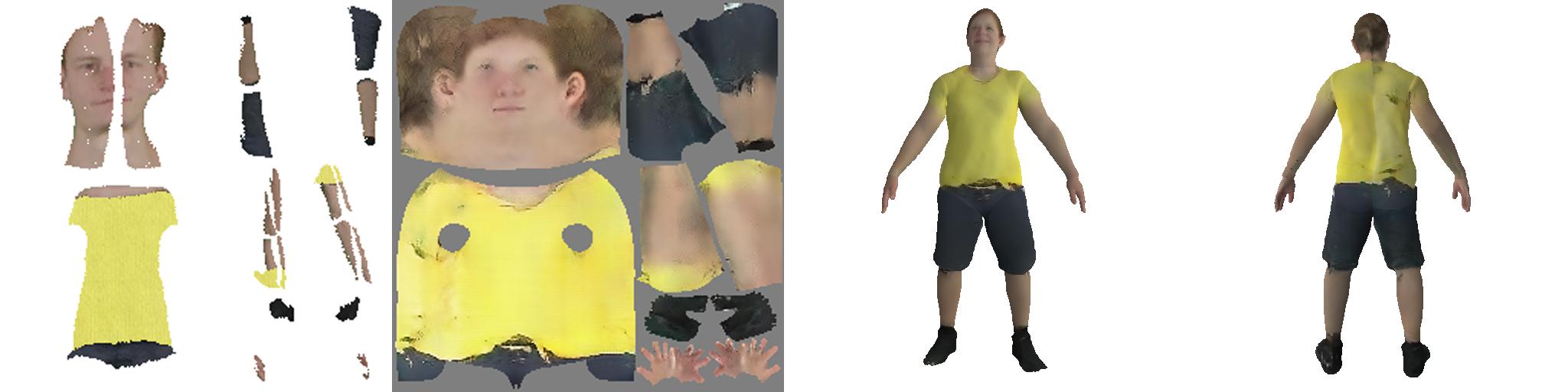}
\includegraphics[width=0.47\textwidth]{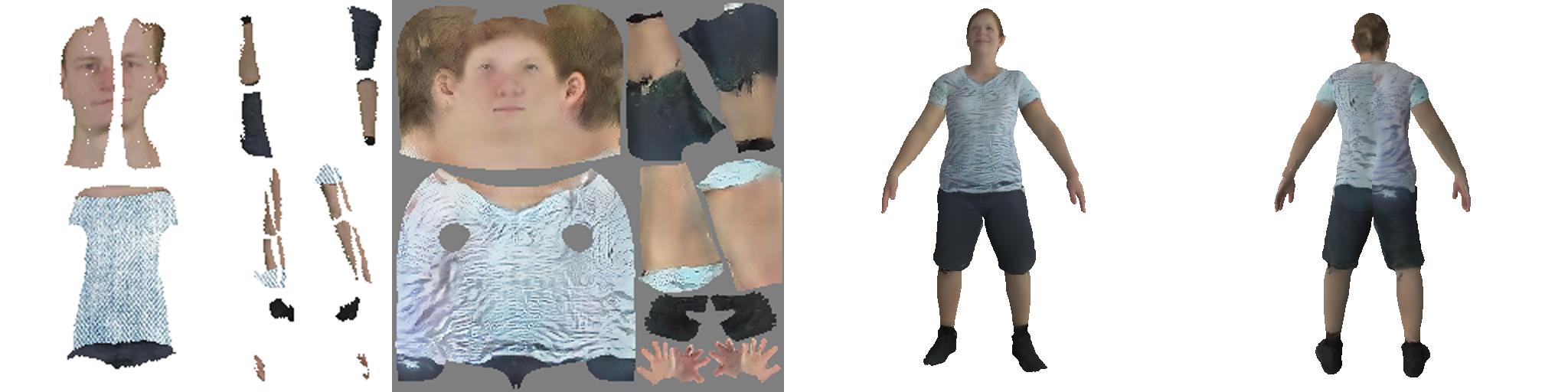}
\caption{Segmentation-based texture editing: the 3D model reconstructed from an image is shown in the top row. The subsequent rows show the edited partial texture and the reconstructed model with different texture for the t-shirt}
\label{fig:texture_edit}
\end{figure}

The full clothing segmentation map gives us the flexibility to edit and adjust the model in different ways. Using the segmentation, one can change the texture of a specific garment and have it completed in a different style. Figure \ref{fig:texture_edit} shows examples where the original T-shirt of the subject is edited with samples of different textures. The network hallucinates wrinkles to make the artificially added texture look more cloth-like.

\begin{figure}[h!]
\centering
\includegraphics[width=0.46\textwidth]{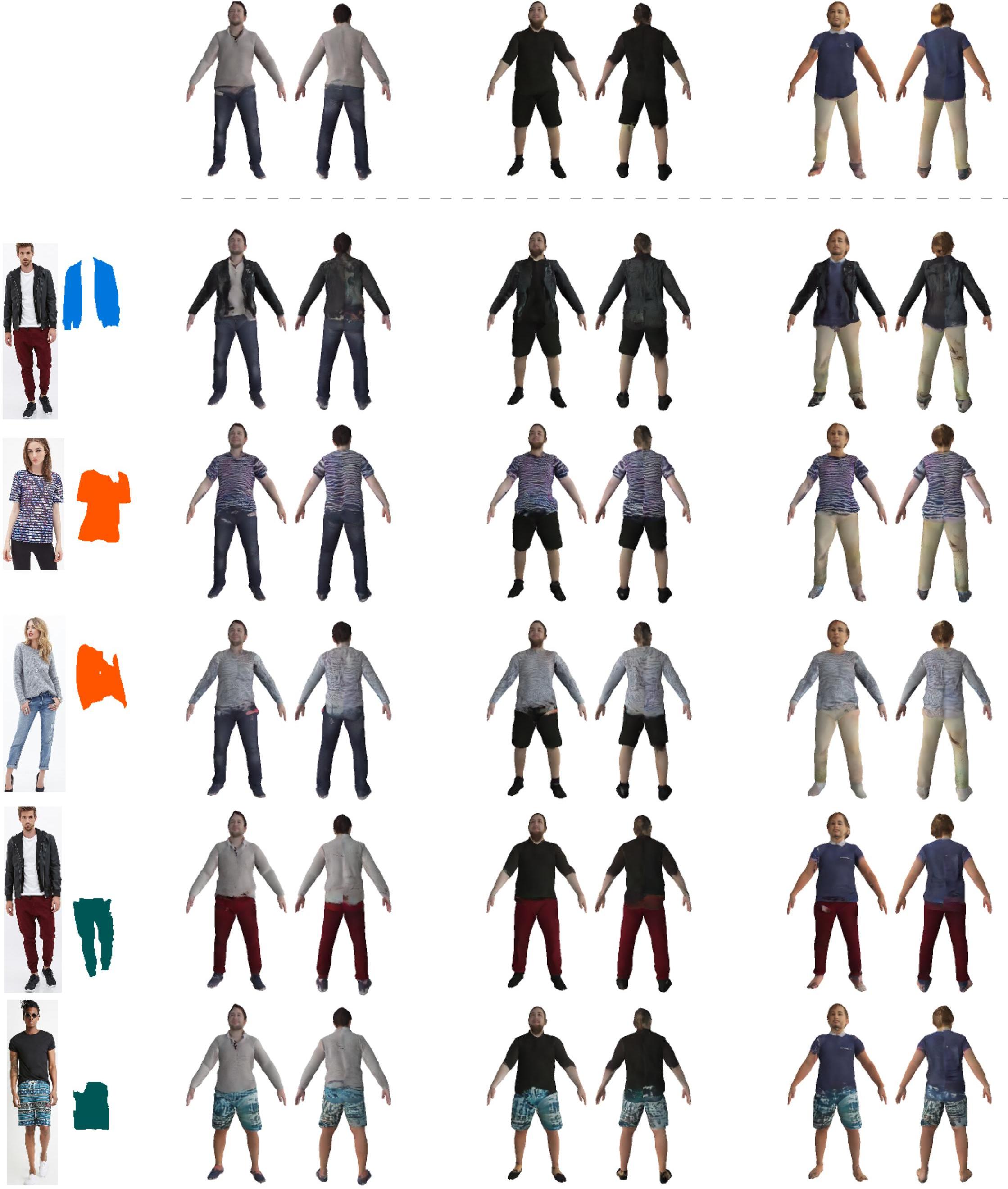}
\caption{Garment re-targeting results: Four subjects are shown in the top row with their 3D  reconstruction. In each of the subsequent rows the subjects are reconstructed wearing the selected garments from the person in the first image.}
\label{fig:redress}
\end{figure}

Additionally, the segmentation could be used for clothes re-targeting. Having two partial texture maps from two images one can rely on the segmentation to exchange garments between the subjects. The edited partial texture and segmentation are then completed, new geometry is produced and the final result is a redressed avatar. An example of this application is shown in Figure~\ref{fig:redress}.

\begin{figure}[h!]
\centering
\includegraphics[width=0.48\textwidth]{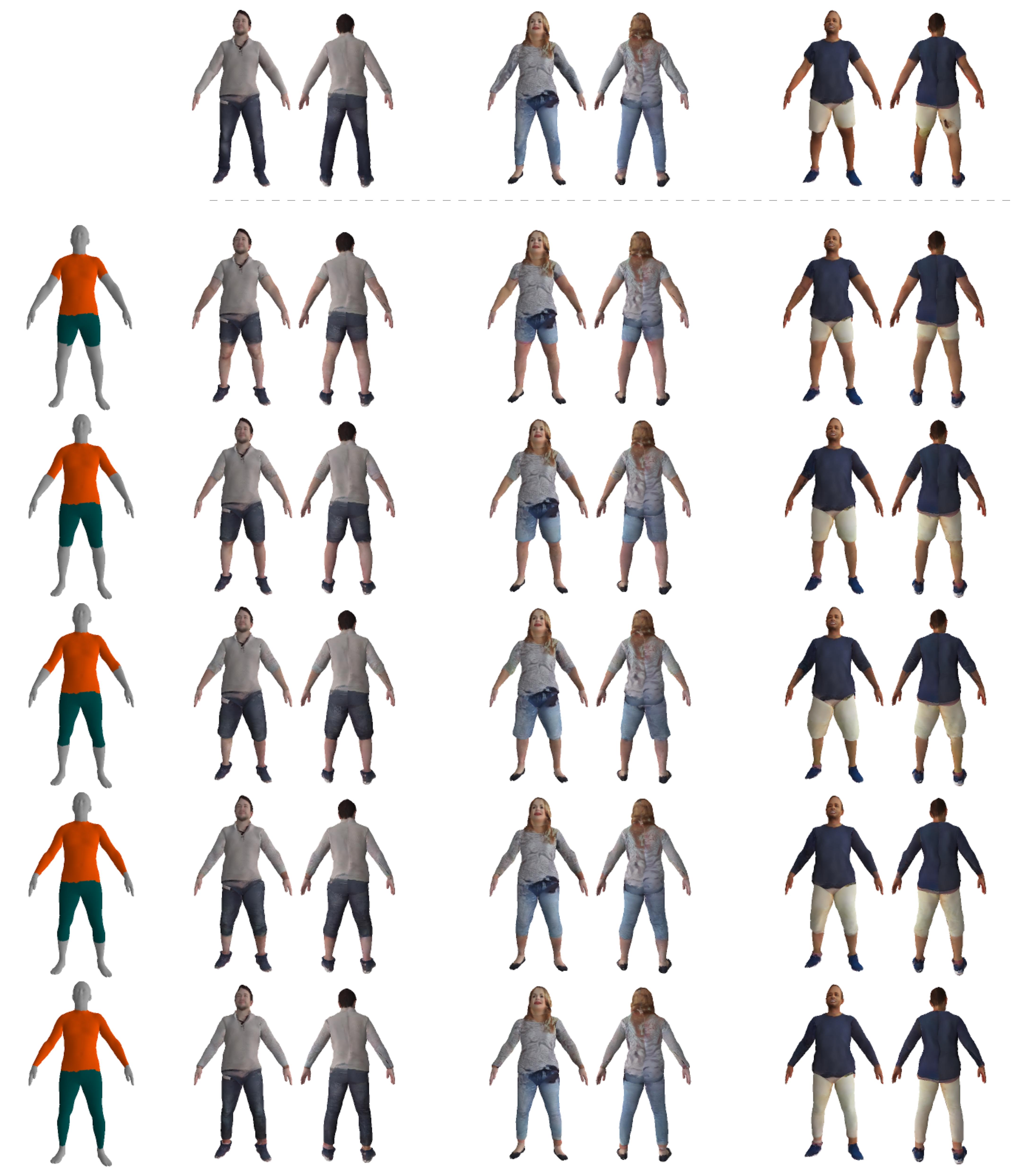}
\caption{Model editing results: Models reconstructed from an image are shown in the top row. In each of the subsequent rows the models are edited such that the length of the sleeves and pants matches the segmentation on the left.}
\label{fig:sleeves_and_pants}
\end{figure}

Finally, one can use the segmentation to edit the length of a garment, transforming shorts to pants or blouse to a T-shirt. We train a separate network that only completes the arms and the legs of the person conditioned on the segmentation. Therefore, once we have reconstructed a model, we can run it through the editing network to change the length of the sleeves or the pants. An example of this scenario is shown in Figure \ref{fig:sleeves_and_pants}. 

For additional results and better preview please check the supplementary material and video \footnote{\href{https://youtu.be/h-HyFq2rYO0}{https://youtu.be/h-HyFq2rYO0}}.
\section{Conclusions and Future Work}
\label{sec:conclusion}
We presented a method for completion of full texture and geometry from single image. Our key idea is to turn a hard 3D inference problem into an image-to-image translation which is amenable to CNNs by encoding appearance, geometry and clothing layout on a common SMPL UV-space. By learning in such space we obtain visually pleasing and realistic 3D model reconstructions.
To allow for control we decompose the predictions into factors: texture maps to control appearance, displacement maps for geometry, segmentation layout to control clothing shape and appearance. We have demonstrated how to swap clothing between 3D avatars just from the image and how to edit the garment geometry and texture. 

There are many avenues for future work. First, our model cannot predict clothing with vastly different topology than the human body, eg. skirts or dresses. Implicit function based representations~\cite{xu2019disn,saito2019pifu,mescheder2018occupancy,natsume2018siclope,huang2018deep} might be beneficial to deal with different topologies, but they do not allow control. Although it is remarkable that our model can predict the occluded appearance of the person, the model struggles to predict high frequency detail and complex texture patterns. We have extensively experimented with style-based losses and models~\cite{gatys2016image,johnson2016perceptual}, but the results are either not photo-realistic, lack control or need to be re-trained per texture. In the future, we would like to investigate models capable of producing photo-realistic textures, and complex clothing without losing control. 

There is a strong trend in the community towards data-driven image-based models for novel viewpoint synthesis~\cite{ma2018disentangled,ma2017pose}, virtual try-on~\cite{han2018viton,wang2018toward}, reposing~\cite{chan2018everybody} and reshaping with impressive results, suggesting 3D modelling can be by-passed. In this paper, we have demonstrated that full 3D completion and 3D reasoning has several advantages, particularly in terms of control and ability to generalize to novel poses and views. 
In the future, we plan to explore hybrid methods that combine powerful 3D representations with neural based rendering~\cite{martin2018lookingood}. \\

\noindent \small {\bf Acknowledgements:} We thank Christoph Lassner, Bernt Schiele and Christian Theobald for provided insight and expertise at the early stages of the project; Thiemo Alldieck and Bharat Lal Bhatnagar for the in-depth discussions and assistance in our dataset preparation. Special thanks should be given to Mohamed Omran for contributing with an abundance of valuable ideas and most honest critiques. 

This work is partly funded by the Deutsche Forschungsgemeinschaft (DFG, German Research Foundation) - 409792180 (Emmy Noether Programme, project: Real Virtual Humans) and the Google Faculty Research Award.

\clearpage
{\small
\bibliographystyle{ieee}
\bibliography{references}
}

\end{document}